%% file: main.tex
\documentclass[letterpaper, 10 pt, journal, twoside]{IEEEtran}
\usepackage{graphics} 
\usepackage{epsfig} 
\usepackage{times} 
\usepackage{amssymb}  
\usepackage[bookmarks=true]{hyperref}
\usepackage{xcolor}
\usepackage{color}
\usepackage[ruled,vlined]{algorithm2e}
\usepackage{siunitx}
\usepackage{colortbl}
\usepackage{multirow}
\usepackage{csquotes}
\usepackage{gensymb}

\newcommand{\bs}{\boldsymbol}

\newcommand{\gameAI}{{AI}}

\usepackage{cite}

\ifCLASSINFOpdf
\else
\fi
\usepackage{amsmath}
\usepackage{url}

\begin{document}
%
\title{Super-Human Performance in Gran Turismo Sport \\
Using Deep Reinforcement Learning}
%
%
%

\author{Florian Fuchs$^{1}$, Yunlong Song$^{2}$, Elia Kaufmann$^{2}$, Davide Scaramuzza$^{2}$, and Peter D\"urr$^{1}$%
\thanks{Manuscript received: October 15, 2020; Revised January 9, 2021; Accepted February 12, 2021.}
\thanks{This paper was recommended for publication by Markus Vincze upon evaluation of the Associate Editor and Reviewers' comments.
This work was supported by Sony R\&D Center Europe Stuttgart Laboratory 1, the National Centre of Competence in Research (NCCR) Robotics through the Swiss National Science Foundation, and the SNSF-ERC Starting Grant.} 
\thanks{$^{1}$F. Fuchs and P. D\"urr are with Sony Europe B.V., Schlieren/Switzerland Branch
        {\tt\footnotesize (Florian.Fuchs@sony.com)}}%
\thanks{$^{2}$Y. Song, E. Kaufmann, D. Scaramuzza are with the Robotics and Perception Group, University of Zurich, Switzerland
        {\tt\footnotesize (\protect\url{http://rpg.ifi.uzh.ch})}}%
\thanks{Digital Object Identifier 10.1109/LRA.2021.3064284}
}
%
%

\markboth{IEEE Robotics and Automation Letters. Preprint Version. Accepted February, 2021}
{Fuchs \MakeLowercase{\textit{et al.}}: Super-Human Performance in Gran Turismo Sport using Deep Reinforcement Learning} 

%



\maketitle

\input{sections/abstract}


\section*{Supplementary Videos}
This paper is accompanied by a narrated video of the performance: \url{https://youtu.be/Zeyv1bN9v4A}\\

%
\IEEEpeerreviewmaketitle

\input{sections/introduction.tex}
\input{sections/relatedwork.tex}
\input{sections/methodology.tex}
\input{sections/experiments.tex}

\input{sections/results.tex}
\input{sections/conclusion.tex}

\section*{Acknowledgment}
We are very grateful to Polyphony Digital Inc. for enabling this research.
Furthermore, we would like to thank Kenta Kawamoto and Takuma Seno from Sony R\&D Center Tokyo for their kind help and many fruitful discussions.

\ifCLASSOPTIONcaptionsoff
  \newpage
\fi



%
\bibliographystyle{IEEEtran}
\bibliography{ref}

\clearpage
\input{sections/supplementary.tex}
%

\end{document}

%% file: sections/abstract.tex
\begin{abstract}
Autonomous car racing is a major challenge in robotics. 
It raises fundamental problems for classical approaches such as planning minimum-time trajectories under uncertain dynamics and controlling the car at the limits of its handling. 
Besides, the requirement of minimizing the lap time, which is a sparse objective, and the difficulty of collecting training data from human experts have also hindered researchers from directly applying learning-based approaches to solve the problem. 
In the present work, we propose a learning-based system for autonomous car racing by leveraging a high-fidelity physical car simulation, a course-progress proxy reward, and deep reinforcement learning. 
We deploy our system in Gran Turismo Sport, a world-leading car simulator known for its realistic physics simulation of different race cars and tracks, which is even used to recruit human race car drivers.
Our trained policy achieves autonomous racing performance that goes beyond what 
had been achieved so far by the built-in AI, and, at the same time, outperforms
the fastest driver in a dataset of over 50,000 human players.
\end{abstract}

%% file: sections/introduction.tex
\section{Introduction}
\begin{figure}[!htp]
    \centering
    \includegraphics[width=\linewidth]{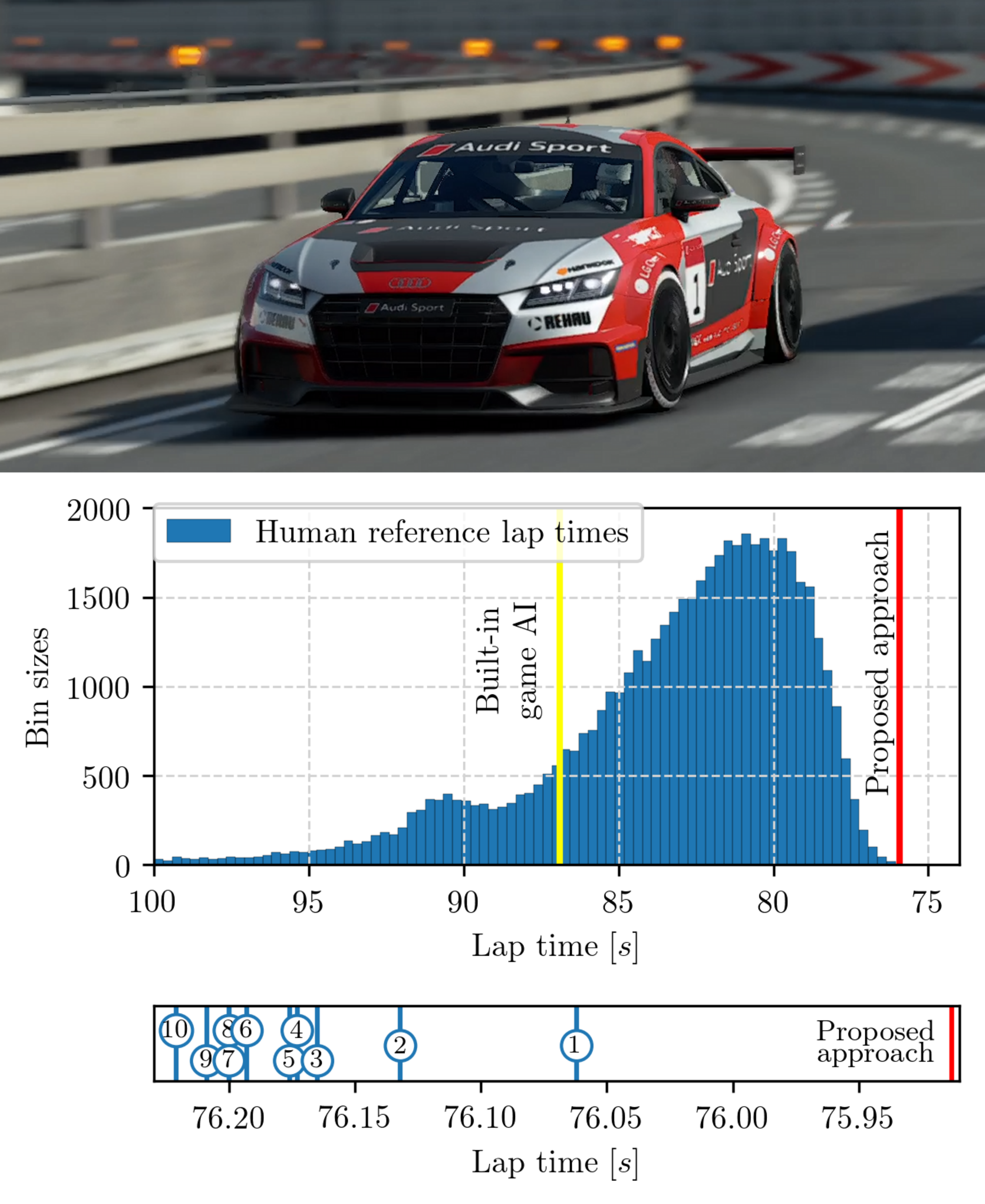}
    \caption{\textbf{Top:} Our approach controlling the ``Audi TT Cup" in the \textit{Gran Turismo Sport} simulation.
    \textbf{Center:} Time trial lap time comparison between our approach (red line), the personal best lap times of over 50,000 human drivers from 70 countries (dark blue histogram), as well as the built-in non-player character (``game AI", yellow line).
    Human lap times over 100 seconds are cut off to simplify the visualization.
    \textbf{Bottom:} Closer view at the lap times of the 10 fastest human drivers and our approach.
    \textbf{Note:} Starting with GTS update 1.57 the simulation dynamics changed such that the achievable times for the tested settings are different. See appendix \ref{sec:reproducibility} for updated results under the new dynamics.}
     \label{fig:lap_time_comparison_tt_cup}
  \end{figure}
  
\IEEEPARstart{T}{he} goal of autonomous car racing is to complete a given track as fast as possible, 
which requires the agent to generate fast and precise actions even when 
the vehicle reaches its physical limits. 
Classical approaches~\cite{aggress_mppi, opt_based_alex, frazzoli2002real} generally decouple the racing problem 
into trajectory planning and vehicle control. 
From the trajectory planning standpoint, the problem is about how to compute the minimum-time 
trajectory for a given race track.
From the vehicle control standpoint, it captures the main difficulty and the most critical scenario in autonomous driving---how to safely control the vehicle during very extreme manoeuvres.
\IEEEpubidadjcol

Classical approaches have been widely studied for addressing the autonomous car racing problem
and have shown impressive results. 
In particular, much of the success relies on optimization-based techniques for trajectory generation and tracking,
thanks to their capabilities of handling different constraints and nonlinear dynamics. 
However, several limitations exist in this line of research, such as the requirement of expensive computations
for the nonlinear optimization and the lack of flexibility in the objective (most optimizations require quadratic cost formulations).
Overall, conventional systems for autonomous driving escalate in complexity as the model complexity increases and 
do not leverage the large volume of data generated by real-world driving.

The requirements of improving vehicle models using real-world data and handling more complex cost formulations and 
general nonlinear dynamics motivated researchers to design learning-based systems. 
The first is related to dynamic model identification. 
In particular, multi-layer neural network models have shown to be helpful for high-speed driving when 
combining them with model predictive path integral control~(MPPI)~\cite{aggress_mppi, info_mpc, williams2018information} or a simple feedforward-feedback control~\cite{spielberg2019neural}.
However, in order to run MPPI online in a receding horizon manner in a fast control loop, the optimization
requires a highly parallel sampling scheme and heavily relies on graphics processing units (GPUs).

The second trend pertains to end-to-end vehicle control using neural networks, where 
a trained neural network policy can map high-dimensional raw observations directly to control commands. 
Hence, it has the advantage of forgoing the need for explicit state estimation and solving optimizations online,
rendering fast and adaptive control performance.
Combined with advanced dynamic model identification, neural network controllers
have shown numerous achievements in many real-world robot 
applications~\cite{hwangbo2019learning, lee2020learning, hwangbo2017control, nagabandi2020deep}.  

In this work, we tackle the autonomous car racing problem in Gran Turismo Sport (GTS), 
which is a world-leading car simulator known for its realistic modeling of various race cars and tracks. 
GTS is a unique platform for studying autonomous racing systems, as it allows
simulating a large number of realistic environments and benchmarking comparisons between autonomous systems 
and experienced human drivers. 
Such comparisons can offer invaluable insights into vehicle control during 
extreme manoeuvres. 

Our approach makes use of model-free deep reinforcement learning and a 
course-progress proxy reward to train a multilayer perceptron 
policy for the vehicle control.
The key is formulating the minimum-time racing problem properly into maximizing the course progress
and choosing meaningful low-dimensional state representations for efficient policy training. 
As a result, we demonstrate the first super-human control performance on the task of 
autonomous car racing.
The strength of our controller is two-fold: 
1) it does not rely on high-level trajectory planning and following
(which human drivers tend to do)
2) it generates trajectories qualitatively similar to those chosen by the best human drivers while
outperforming the best known human lap times.
Our findings suggest that reinforcement learning and neural network controllers merit further
investigation for controlling autonomous vehicles at their limits of handling.

%% file: sections/relatedwork.tex
\section{Related Work}

Prior work in the domain of autonomous racing can be grouped into three groups: (i) classical approaches relying on trajectory planning and following, (ii) supervised learning approaches, and (iii) reinforcement learning approaches. 

\textbf{Classical Approaches}
Classical approaches to autonomous car racing approach the problem by separating it in a chain of submodules consisting of perception, trajectory planning, and control.
In particular, model predictive control (MPC)~\cite{opt_based_alex, novi2019real, rosolia2019learning, kabzan2019amz, ostafew2016robust, verschueren2014towards} is a promising approach for controlling the vehicle at high speed.
In~\cite{kabzan2019learning}, an MPC controller is combined with learned system dynamics based on Gaussian Processes for the task of autonomous car racing. While being able to capture the complex dynamics of the car, the resulting controller requires trading-off dynamical fidelity against computation time.
Similarly, model predictive path integral control (MPPI)~\cite{aggress_mppi, info_mpc, williams2018information} is a more flexible approach that can be combined with complex cost formulations and neural network vehicle models. 
Both MPC and MPPI have shown impressive results in controlling physical vehicles at high speed in the real-world.
Despite their successes, both approaches have limitations, such as the lack of flexibility in the cost function design 
or the requirement of highly parallel computations. 
Moreover, hierarchical approaches are also susceptible to failures of each submodule, 
where the operational states of each module are limited by a series of approximations and linearizations.

\textbf{Imitation Learning}
Instead of planning trajectories and tracking them with a controller, imitation-based approaches directly learn a mapping from observation to control action in a supervised fashion. Learning such mapping requires labelled data, which is typically provided by human expert demonstrations or a classical planning and control pipeline. For example, ALVINN~(Autonomous Land Vehicle in a Neural Network)~\cite{pomerleau1989alvinn} is one of the first autonomous driving systems that uses a neural network to follow a road.
Similarly, a convolutional neural network~(CNN) controller was trained by~\cite{bojarski2016end} for lane and road following.
Pan et al.~\cite{pan2018agile} use imitation learning for agile, off-road autonomous driving.
Distilling human or algorithmic expert data in a learned policy is a promising approach to overcome the strict real-time constraint of classical approaches, but its performance is, by design, upper-bounded by the quality of the training data.

\textbf{Reinforcement Learning}
Model-free reinforcement learning optimizes parametrized policies directly based on sampled trajectories, and hence, does not suffer from issues that hamper the aforementioned approaches, such as the necessity of solving nonlinear optimizations online or the dependence on labelled training data.
For example, a number of studies~\cite{ete_race_driving, drive_in_20m, drive_in_1d, grigorescu2020survey, cai2020high} have demonstrated the success of using model-free deep RL for end-to-end driving and racing. 
Recently, a high-speed autonomous drifting system~\cite{cai2020high} was developed in simulation 
using the soft actor-critic~(SAC) algorithm~\cite{haarnoja2018soft}.
Here, off-policy training plays an important role in~\cite{drive_in_20m, drive_in_1d, cai2020high}, in which the high sample complexity previously limiting the widespread adoption of deep RL methods in high-dimensional domains, was substantially reduced by combining offline training with an experience replay buffer.
Despite the successful application of deep RL algorithms in real as well as simulated autonomous driving, there is, to the best of our knowledge, no work matching or exceeding the performance of human expert drivers in terms of speed.

%% file: sections/methodology.tex
\section{Methodology}
\begin{figure}[!htp]
  \centering
  \includegraphics[width=\linewidth]{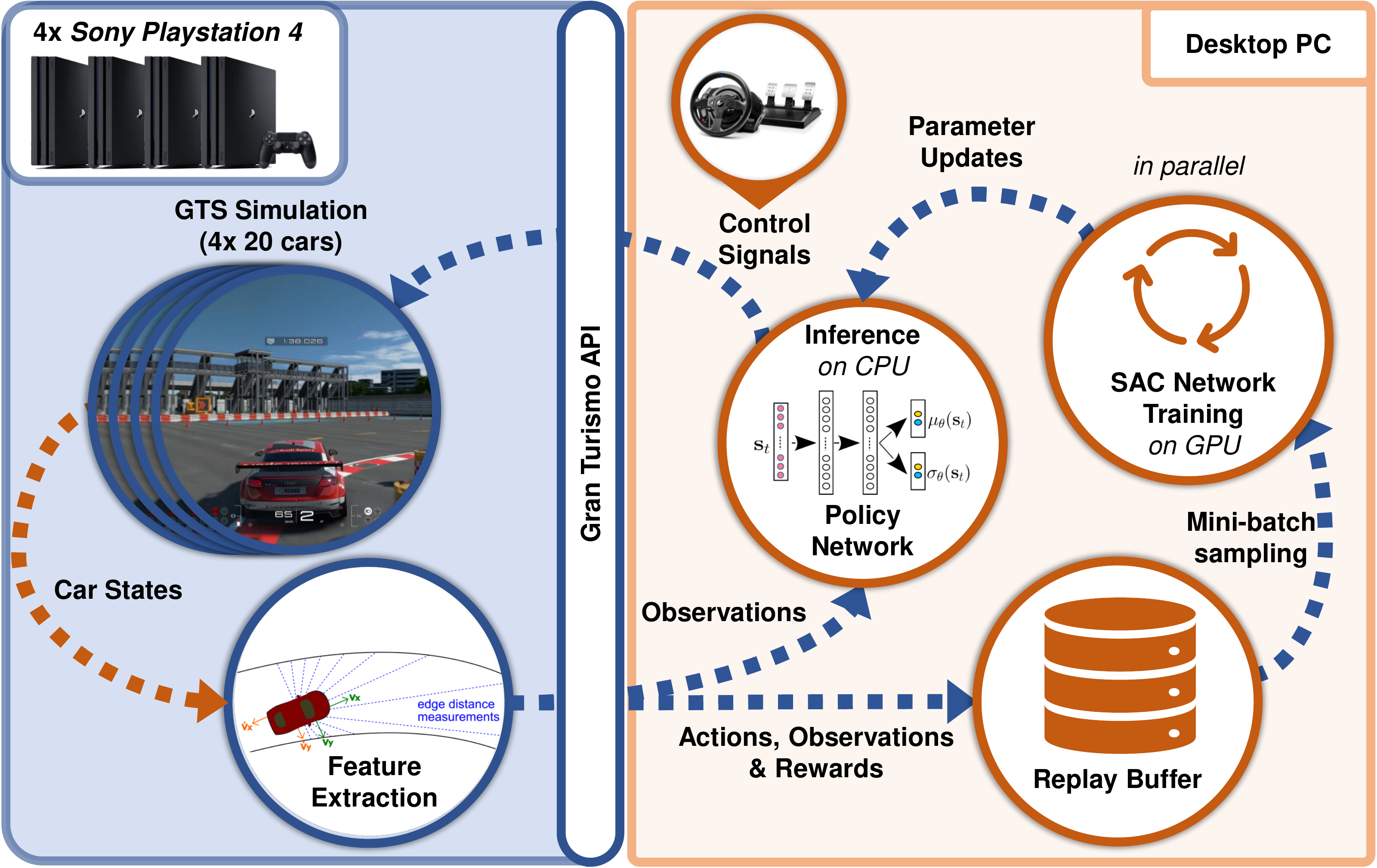}
  \caption{System overview: we train a policy network to directly map observations, including a set of range finder measurements and the car's velocity and acceleration, to control commands consisting of the car's steering angle as well as the level of throttle and brake.
  We use a distributed sampling scheme to collect samples from 4 \textit{PlayStation~4} simulating 20 cars each, and then,
  store the sampled trajectory in a fixed size first in, first out (FIFO) replay buffer.
  In parallel, we optimize the policy parameters using a soft actor-critic algorithm and data uniformly 
  sampled from the replay buffer.}
  \label{fig:sys_diag}
\end{figure}

Our main goal is to build a neural network controller that is capable of autonomously navigating a race car without 
prior knowledge about the car's dynamics while minimizing the traveling time on a given track in the GTS environment. 
To achieve this goal, we first define a reward function that formulates the racing problem and a neural
network policy that maps input states to actions.
We then optimize the policy parameters 
by maximizing the reward function using the SAC~\cite{haarnoja2018soft} algorithm.
An overview of our system is shown in Figure~\ref{fig:sys_diag}.
%

\subsection{Minimum-Time Problem and Reward Function}\label{sec:reward}
\begin{figure}[!ht]
     \centering
     \includegraphics[width=0.95\linewidth]{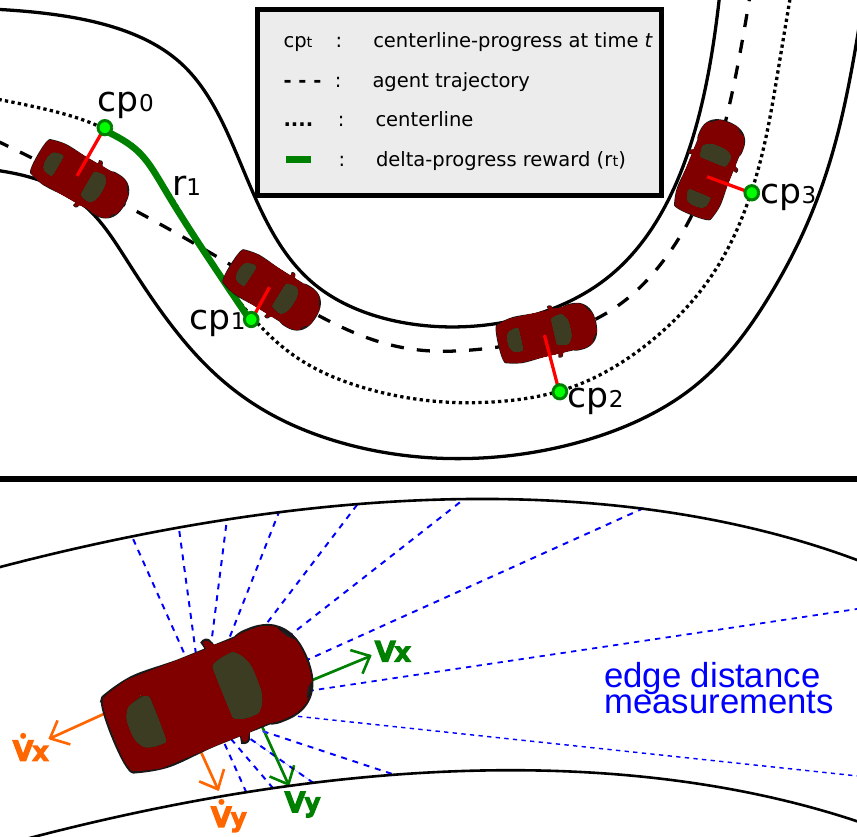}
     \caption{\textbf{Top:} The course progress $cp_t$ at time $t$ is constructed by projecting the car's position on the track's center line.
     We then define the progress-based reward $r^\text{prog}$ as the increment in centerline-progress $r^\text{prog}_{t} = cp_{t} - cp_{t-1}$.
     \textbf{Bottom:} A subset of the observations fed to the policy and value networks.}
     \label{fig:reward_state}
    \end{figure}

    We aim to find a policy that minimizes the total lap time for a given car and track.
    The lap time itself is a very sparse reward signal.
    Changes in the signal are therefore hard to attribute to specific actions of an agent.
    We design a proxy reward based on the current course progress, which can be evaluated in arbitrary time intervals.
    To reduce the variance of the reward signal, we use an exponential discount for future rewards, with discount factor $\gamma$.
    Maximizing the course progress closely approximates minimizing the lap time when choosing a high enough discount factor.
    The new proxy reward allows trading-off an easily attributable but biased reward and a reward closer to the overall lap time goal by adjusting $\gamma$, and thus changing the weight of future rewards based on their temporal distance to an action and state.
    The construction of the progress reward can be seen in the top of Figure~\ref{fig:reward_state}.
    
    The use of an exponentially discounted future reward leads to a bias towards short-term rewards.
    In the GTS setting this bias reduces the incentive for an RL agent to brake, e.g. to prevent crashes.
    To counteract this short-term bias, we introduce a second reward term that penalizes wall contact relative 
    to the car's kinetic energy, leading to the final reward function of
    \begin{equation}\label{eq:reward_total}
    r_t = r^\text{prog}_{t} -
    \left\{\begin{matrix}
      & c_w \lVert\mathbf{v}_{t}\rVert^{2} & \text{if in contact with wall,}  \\
      & \quad 0 & \text{otherwise}
    \end{matrix}\right.
    \end{equation}
    where $c_{w} \geq 0$ is a hyperparameter controlling the trade off between wall contact penalty and progress-based reward. 
    Here $\mathbf{v}_{t}=[v_x, v_y, v_z]$ is a velocity vector that represents the current 
    linear velocity of the vehicle.
    A similar approach to incentivize collision avoidance was introduced in~\cite{kahn2017uncertainty}.
    The introduction of the kinetic energy is justified through the energy-dependent loss in
    acceleration that takes place when hitting a wall.
    Without this additional wall contact penalty, we found the learned policies did not brake and simply grinded along the track's walls in sharp curves.
    When using fixed valued wall contact penalties, we found the agent either did not react to the penalty or ended up in a strategy of full braking and standing still to not risk any wall contact, depending on the strength of the penalty.

\subsection{Policy Network}

We represent the driving policy with a deep neural network. We make use of the SAC network architecture proposed in the original paper~\cite{haarnoja2018soft}. We use a policy network, two Q-function networks, and a state-value function network, each with 2 hidden layers with 256 ReLU nodes, resulting in a total of 599,566 trainable parameters. In the following, the features fed to the network and its prediction are explained in detail. 

\textbf{Input features} We base the input feature selection on the results of a pre-study on GTS conducted with behavioral cloning, aiming to learn human-like driving based on a regression of experts' actions on visited states.
We iterated over combinations of subsets of features provided by GTS as well as additionally constructed features, which lead to the following feature selection:

1)~The linear velocity $\bs{v}_t \in \mathbb{R}^3$ and the linear acceleration $\bs{\dot{v}}_t \in \mathbb{R}^3$.
2)~The Euler angle $\theta_t \in (-\pi, \pi]$ between the 2D vector that defines the agent's rotation in the horizontal plane
and the unit tangent vector that is tangent to the centerline at the projection point.
This angle is the only direct way for the policy network to detect if a car is facing the wrong direction. 
3)~Distance measurements $\bs{d}_t \in \mathbb{R}^{M}$ of $M$ rangefinders with a maximum range of $\SI{100}{\meter}$ that measure the distance 
from the vehicle's center point to the $M$ edge points of its surrounding objects, such as the edge of the race track.
The rangefinders are equally distributed in the front 180$\degree$ of the car's view.
4)~The previous steering command $\delta_{t-1}$. 
5)~A binary flag with $w_t=1$ indicating wall contact
and 6)~$N$ sampled curvature measurement of the course centerline in the near future $\bs{c}_t \in \mathbb{R}^{N}$.
The curvature is represented through an interpolation of the inverse radii of circles fitted through centerline points provided by GTS.

We therefore represent the observation at a given time step $t$ as a vector denoted as
$\bs{s}_t = [\bs{v}_t, \bs{\dot{v}}_t, \theta_t, \bs{d}_t, \delta_{t-1}, w_t, \bs{c}_t]$.
A subset of the features is illustrated in the bottom of Figure~\ref{fig:reward_state}.
To allow for a fair comparison with human drivers, we only use features that humans can either directly perceive or deduce from the GTS simulation.

\textbf{Network prediction} The output of the policy network $\bs{a}_t = [\delta_t, \omega_t]$ directly encodes the steering angle $\delta_t \in [-\pi/6, \pi/6] \SI{}{\radian}$ and a combined throttle-brake signal $\omega_t \in [-1, 1]$, where $\omega_t = 1$ denotes full throttle and $\omega_t = -1$ full braking. The combination of throttle and brake in a single signal is motivated by an analysis of human recordings, which revealed that fast strategies do not involve any simultaneous use of throttle and brake.

%% file: sections/experiments.tex
\section{Experiments}\label{sec:experiments}
We evaluate our approach in three race settings, featuring different cars and tracks of varying difficulty. We compare our approach to the built-in \gameAI~as well as a set of over 50,000 human drivers. To ensure a fair comparison to the human drivers, we constrain the policy to produce only actions that would also be feasible with a physical steering wheel. In the following, each race setting as well as the constraints imposed for the comparison with human drivers are explained.

\begin{figure}[!htp]
 \centering
 \includegraphics[width=\linewidth]{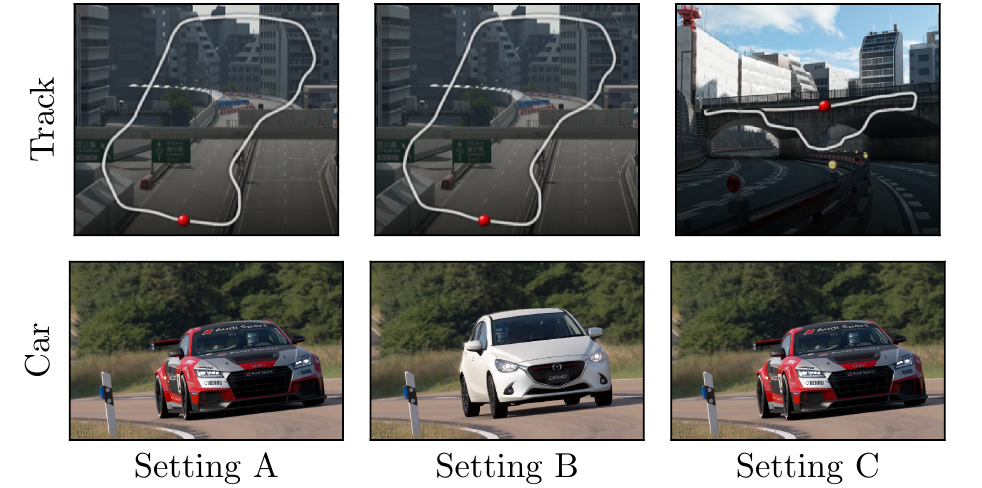}
 \caption{The tracks and cars used as reference settings to compare our approach to human drivers.}
 \label{fig:settings}
\end{figure}

\textbf{Race settings} We train separate agents for three experimental conditions for which human data from past online time trial competitions is available.
The data, provided by GTS manufacturer \textit{Polyphony Digital Inc.}, includes the personal best lap time and trajectory of every competitor, ranging from absolute beginners to World Cup contenders.
The competitions were restricted to fixed settings such as car, course, tires, and racing assistance settings.
This allows for equal conditions when comparing human results with our approach.
Figure~\ref{fig:settings} shows the used cars and tracks.
Setting \textit{A} and \textit{B} use a track layout featuring a typical combination of curves and straight segments. 
The two settings differ in the cars used, with the ``Audi TT Cup '16" of setting \textit{A} having a higher maximum speed, more tire grip, and a faster acceleration than the ``Mazda Demio XD Turing '15" of setting \textit{B}.
Setting \textit{C} features the same car as setting \textit{A} but a more challenging track layout which features a larger range of speeds due to the combination of long straight segments and very tight turns.

\textbf{Leveling the playing field} Whereas human players are limited by the physical inertia of their game controllers, 
our agent can, in principle, make arbitrarily large action changes. 
To allow for a fair comparison between our agent and human players, 
we estimated the maximal action change that a human can achieve within one frame 
with a \textit{Thrustmaster T300} steering wheel and pedals.
Based on that, we restricted the maximum change between frames during evaluation to $\SI{0.03}{\radian}$ for the steering angle and 80\% for each the throttle and brake range\footnote{A similar fairness restriction was put on the agent proposed in~\cite{vinyals2019grandmaster}, by limiting the number of actions an agent could take per minute.}.
We however found this restriction to not significantly influence the resulting lap times.

\textbf{Robustness}
To detect potential shortcoming of our approach for a more general driving task, such as driving on multiple tracks or driving in the real world, we evaluate our agent's robustness by testing its ability to generalize to a range of modified environments.
We deploy the fastest agent trained in setting \textit{A} on the following problem settings without any retraining:
1) transferring the policy to a different car (setting \textit{B}), 2) transferring the policy to a different track (setting \textit{C}), 3) changing the tire friction, 4) adding uniform noise to the agent's observations, and 5) delaying the agent's inference.

%% file: sections/results.tex
\section{Results}
In this section we evaluate our approach on the three reference settings introduced in Section~\ref{sec:experiments}. We compare the lap times obtained by our approach with the ones achieved by AI built into GTS and the fastest human drivers. Additionally, we analyze the resulting driving behaviour and compare it with driving strategies of professional human drivers. Due to the dynamic nature of our experiments, we encourage the reader to watch the supplementary videos.

\subsection{Lap Time Comparison}
Our approach outperforms the best human lap time in all three reference settings, overcoming the limitations of the currently built-in \gameAI, which itself is outperformed by a majority of players.
Table~\ref{tab:daily_race_results} shows the fastest achieved lap times for the three race settings.
    \input{tables/daily_race_table.tex}
On the first track our approach undercuts the fastest human lap times by 0.15 and 0.04 seconds for settings~\textit{A} and \textit{B} respectively. 
We believe that the difference in margins to the best human drivers results from the speed differences between the two used cars, 
with the faster ``Audi TT Cup" in setting~\textit{A} requiring a more agile strategy than the relatively slow ``Mazda Demio" in setting \textit{B}.
While human players possibly struggle with the fast-paced setting, our approach does not suffer under the increased demands.
Due to the similarities of the trajectories of setting~\textit{A} and \textit{B}, in the following part we will only analyze the trajectories of setting~\textit{A}.
Over 10 consecutive laps in setting~\textit{A} our agent's lap time shows a standard deviation of 15~ms.
A top 1\% driver we invited for our study showed a standard deviation of 480~ms.
In setting~\textit{C} our approach undercuts the best human time by 0.62~seconds. As in the other two settings, the increased margin can be explained by the more challenging track layout and faster car that render setting~\textit{C} the most difficult combination in our experiments. 

\begin{figure}[!t]
     \centering
     \includegraphics[width=\linewidth]{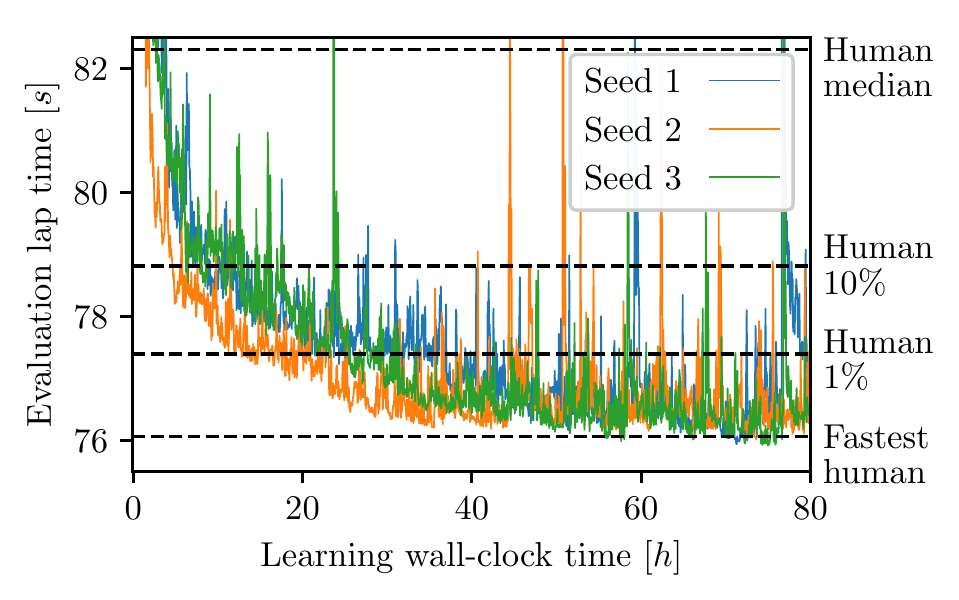}
     \caption{Training progress for our approach learning on setting \textit{A}.
     We evaluate the policy every 2 epochs of learning by letting the agent drive 2 consecutive laps.
     We take the lap time at the second lap as performance metric such that the fixed starting speed applied in the first lap does not distort the result.}
     \label{fig:lap_time_development_tt_cup}
  \end{figure}

Figure~\ref{fig:lap_time_development_tt_cup} shows the learning progress in setting~\textit{A} for three differently initialized neural network policies.
\begin{figure}[!ht]
 \centering
 \includegraphics[width=\linewidth]{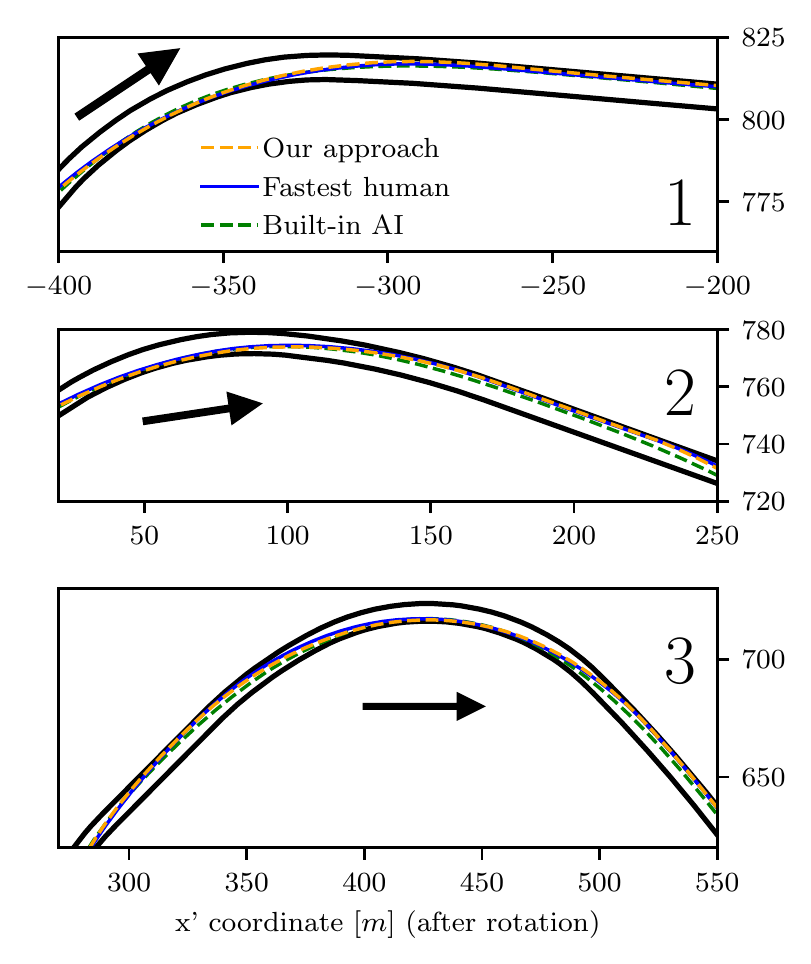}
 \caption{Top-down view of driven paths for setting \textit{A}.
 While the paths of our approach and the human driver are similar, the built-in GTS \gameAI~drives with a safety margin to the track's walls, leading to tighter radii and a reduced speed.}
 \label{fig:zoom_curve}
\end{figure}
\begin{figure*}[!ht]
     \centering
     \includegraphics[width=0.95\textwidth]{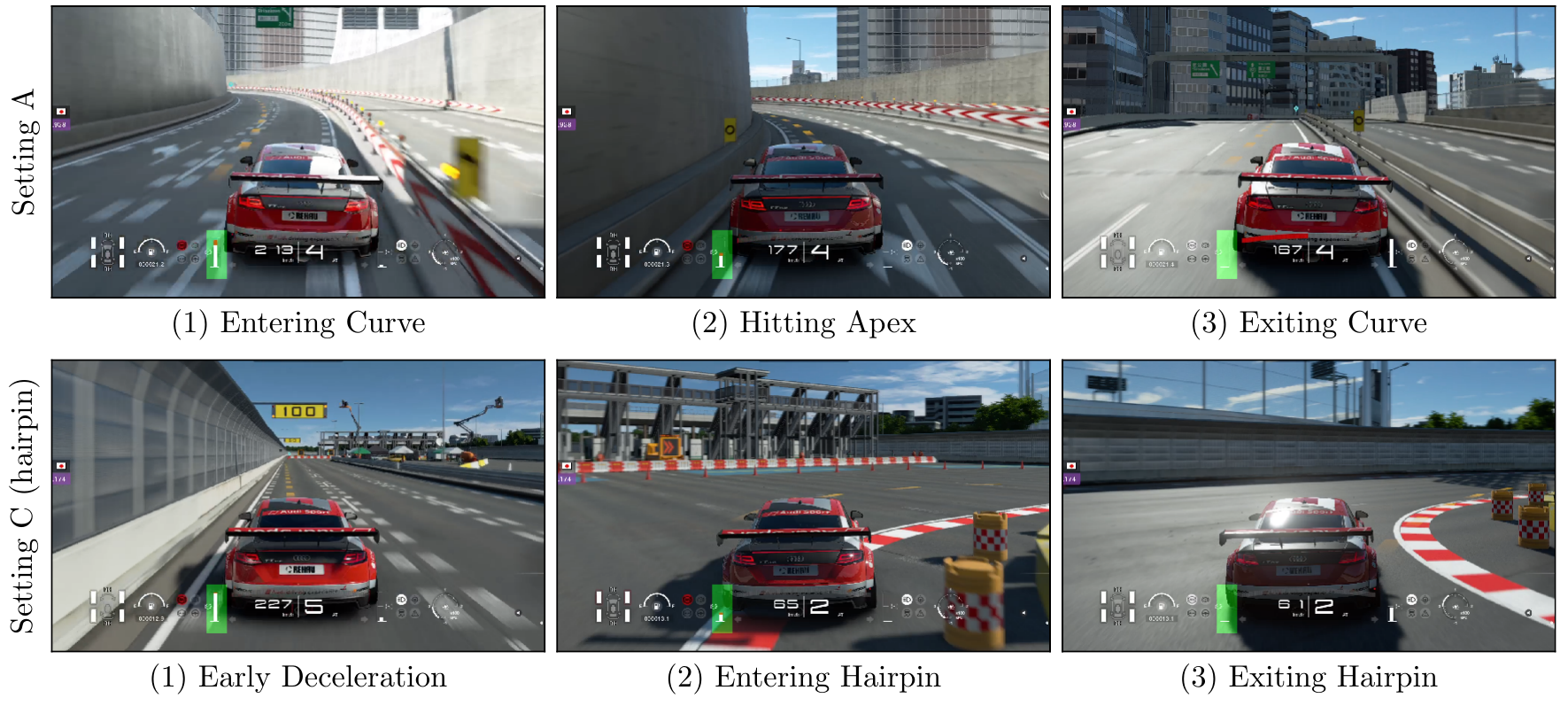}
     \caption{Out-in-out driving behavior and early curve anticipation learned by our approach for setting \textit{A} (top) and setting \textit{C} (bottom). The brake signal is represented by the white bar marked in green. The segments are shown in the supplementary video from 0:24 to 0:30 and 2:50 to 3:04.}
     \label{fig:screenshots}
  \end{figure*}
With each random initialization our approach learns policies that achieve lap times faster than the fastest human reference lap after 56 to 73 hours of training, which on average corresponds to 2,151 training epochs and a total of 946,453 km driven.
The 3 seeds result in similar learning progresses that mainly differ in the time and 
order in which they learn to master certain curves.
Note that warm-starting the training process from a policy that is trained via imitation learning from human data does not improve final performance. 
Training the policy from scratch outperforms the behaviour-cloning policy after less than one hour of training.

\subsection{Learned Driving Behavior}
We analyze the models' driving behavior at the epoch with the fastest evaluation lap time.

\textbf{Out-in-out trajectory} 
Our approach learned to make use of the whole width of the track to maximize its trajectories' curve radii, driving a so-called out-in-out trajectory.
This allows the agent to drive higher speeds before losing traction.
Figure~\ref{fig:zoom_curve} shows that learned behavior in a curve driving comparison between our approach, the fastest human, and the built-in \gameAI~for 3 curves in setting \textit{A}.
Our approach learned to drive curves similar to those of the fastest human, without having access to any human demonstration or using an explicit path planning mechanism.
The top of Figure~\ref{fig:screenshots} shows the learned out-in-out trajectory for the sharpest curve of setting \textit{A}.

\textbf{Anticipating curves}
Furthermore, our approach learned to detect curves early enough and assess their sharpness to decelerate to speeds that allow completing the curves without overshooting into the track's walls, while at the same time not being overly cautious.
This can be seen in the sharpest curve of setting \textit{A} in the top of Figure~\ref{fig:screenshots} and in a more extreme case in the hairpin of setting \textit{C} in the bottom of the Figure.
In the hairpin, the agent starts decelerating $\sim$100 meters before the start of the curve.

\begin{figure}[!htp]
 \centering
 \includegraphics[width=\linewidth]{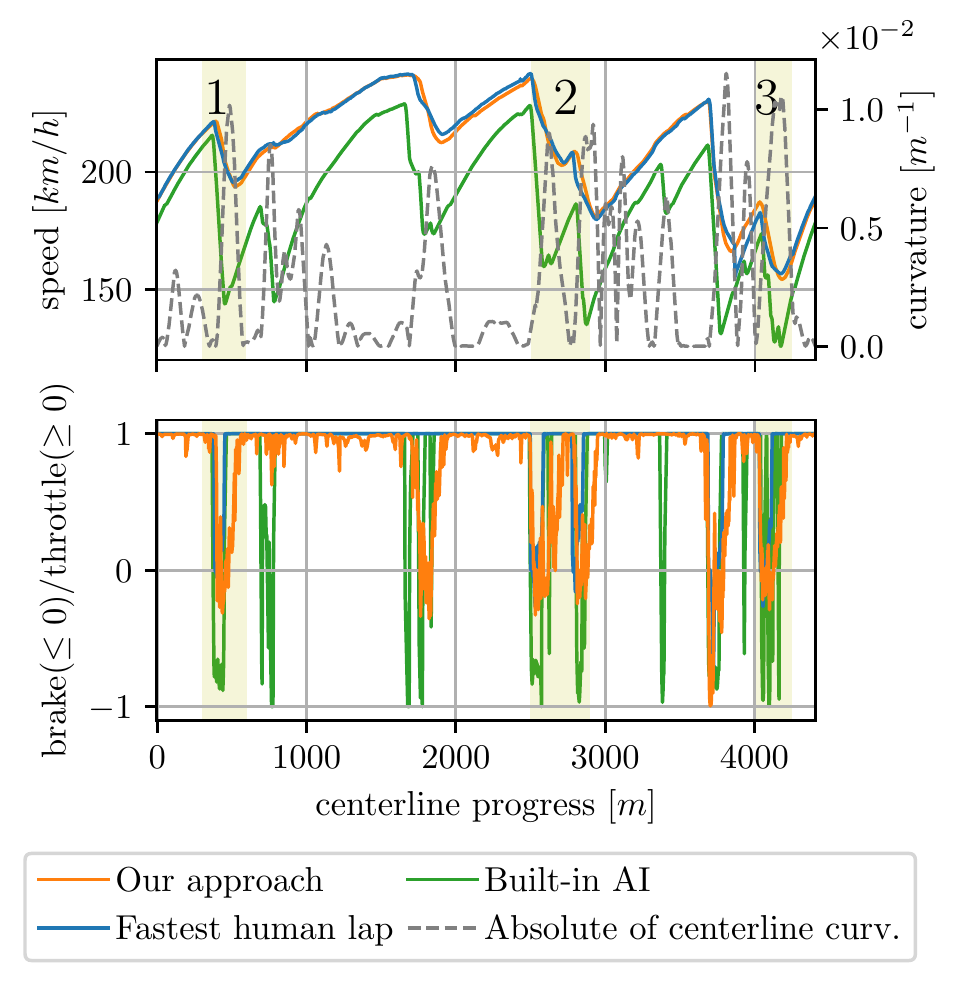}
 \caption{\textbf{Top:} Comparison of speed between the fastest human, our approach, and the built-in GTS AI in setting \textit{A}.
 \textbf{Bottom:} Comparison of combined throttle/brake signal.}
 \label{fig:speed_coursev}
\end{figure}

\textbf{Overall speed} The top of Figure~\ref{fig:speed_coursev} shows a speed comparison between our approach, the fastest human lap, and the built-in AI in setting \textit{A}.
Our approach learned a control policy that closely matches and sometimes even improves on the speed as well as the path of the fastest human reference lap.
Even though the path and speed of our approach and the fastest human are similar for setting \textit{A}, our approach undercuts the fastest human lap time by 0.15 seconds.
While this might not seem like a significant difference, this margin is similar to those between top contenders in real and simulated racing championships.

\begin{figure*}[!ht]
     \centering
     \includegraphics[width=0.9\textwidth]{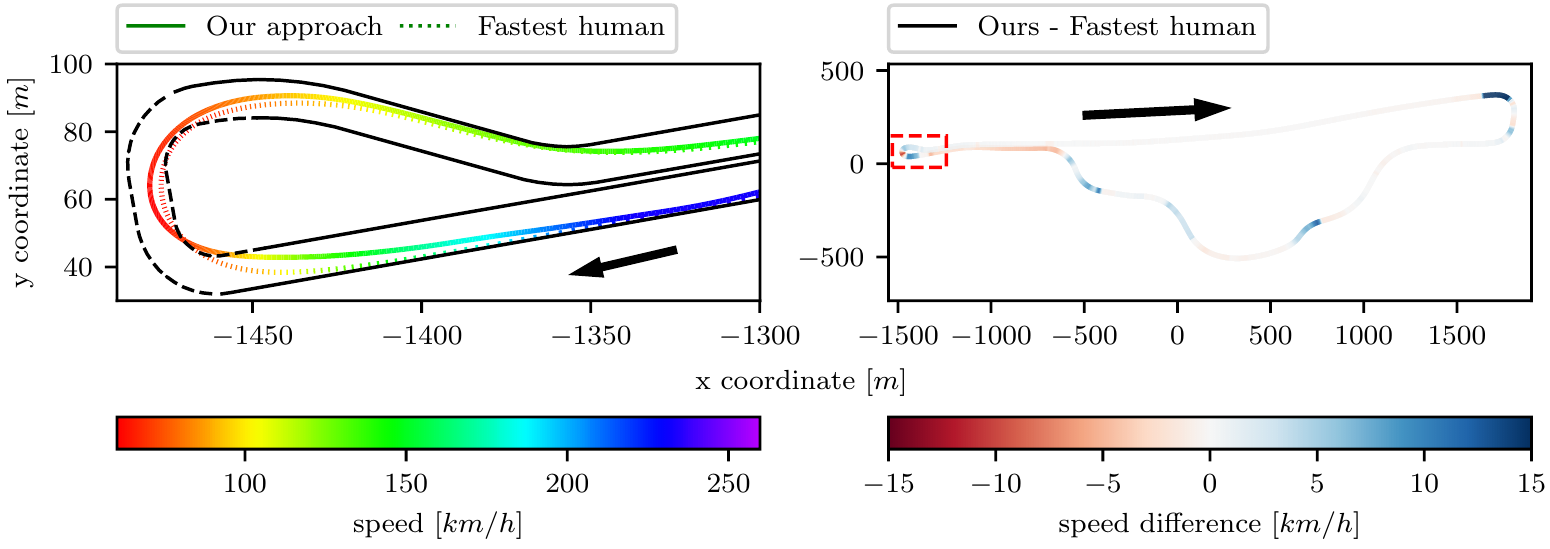}
     \caption{\textbf{Left:} the trajectory driven by our approach and the fastest human trajectory in the hairpin segment in setting \textit{C}. The solid track border is represented by walls on the track, while the dashed track border is simply drawn onto the track.
     To not get off the track, drivers are forced to brake early.
     \textbf{Right:} the speed difference between our approach and the fastest human over the whole track. Positive values indicate our approach driving faster in a segment.}
     \label{fig:hairpin}
\end{figure*}

The left side of Figure~\ref{fig:hairpin} shows the path and speed driven by our approach and the fastest human in the hairpin curve.
While the human expert drives a wider trajectory on the incoming straight, our approach compensates on the outgoing straight, leading to a similar curve exit speed and curve completion time.
The right side of Figure~\ref{fig:hairpin} shows the speed difference between our approach and the fastest human.
Our approach achieves similar or higher speeds over most of the track, often driving tighter and faster curves than the human expert, which leads to an overall faster time. 

In summary, these results indicate that our approach learns to drive autonomously at high-speed in different race settings, using different cars and driving on different tracks, including a challenging hairpin curve.
In all settings our approach achieved lap times faster than those of all human reference drivers.
Moreover, we find that our approach learns to drive trajectories that are qualitatively similar to those chosen 
by the best human players while maintaining slightly higher average speeds in curves by successfully executing later brake points.

\subsection{Robustness}

We analyze the robustness of our trained agent by modifying the test environment.

\textbf{1) Transfer to new car:}
The agent still follows an out-in-out trajectory and is able to finish the track.
In straight segments as well as slight curves, the agent is able to adjust to the change in dynamics and execute a feasible trajectory.
In sharp curves however, the agent sometimes overshoots the track.
This is most likely caused by the training car being easier to control than the test car, due to higher tire-surface friction caused by both the weight and the racing tires of the training car.
For sharper curves, the agent is not able to extrapolate its behavior to the previously unseen dynamics of a car with less tire-surface friction.

\textbf{2) Transfer to new track:}
The agent is able to drive wall-contact free in straight segments as well as slight curves. 
It is however not able to extrapolate its behavior to some of the unseen curve shapes.
Since the agent gets stuck in the hairpin curve, it is not able to finish the track.

\textbf{3) Increased tire friction:}
The agent is able to adjust to the change in friction by making corrections when deviating from the expected trajectory.
In 2 curves, the agent over-estimates the outward force, leading to brief contact with the inner edge of the curve.
This leads to a loss of 0.1~seconds on the baseline.
\textbf{Lowered tire friction:}
The agent is able to make corrections to its trajectory to adjust to the change in dynamics and drive out-in-out paths for all but the 3 sharpest curves.
In the sharpest curves, the agent takes a too aggressive path, based on the higher friction experienced in training, leading to contact with the outer edge at the end of the curve.

The two experiments indicate that our agent is able to make corrections to its path to a certain degree.
However, for some track segments the corrections are not strong enough to handle the change in dynamics.

\begin{figure}[!ht]
     \centering
     \includegraphics[width=0.9\linewidth]{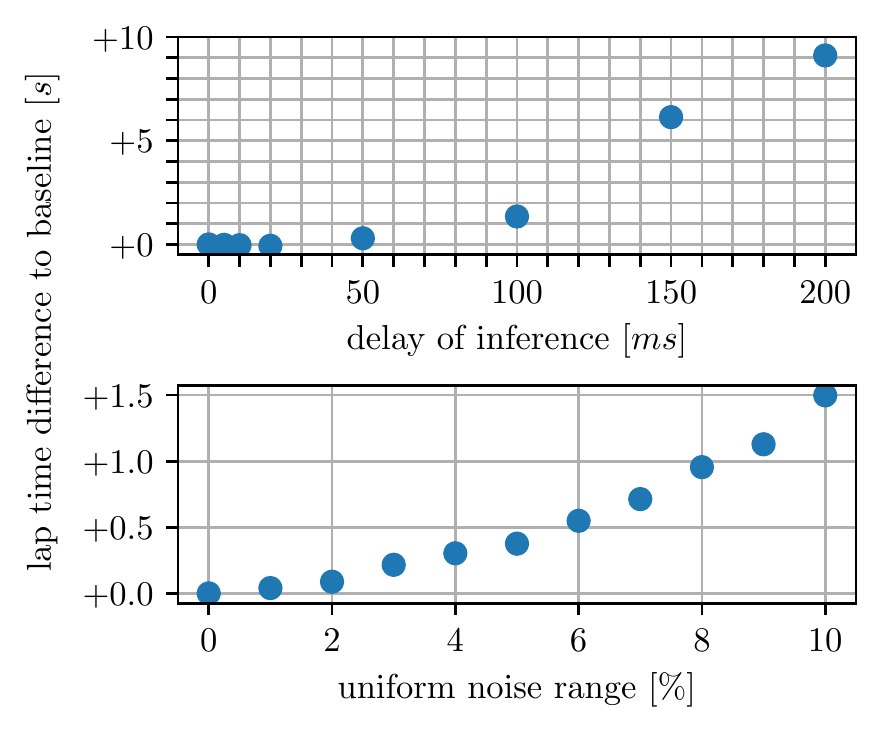}
     \caption{\textbf{Top:} Loss in lap time when artificially delaying the agent's inference compared to undelayed baseline agent in setting \textit{A}. \textbf{Bottom:} Loss in lap time when adding noise to the agent's observations.}
     \label{fig:delay_noise_vs_laptime}
\end{figure}

\textbf{4) Noise in Observations:}
The bottom of Figure~\ref{fig:delay_noise_vs_laptime} shows the lap times in setting \textit{A} for different degrees of uniform noise added to the agent's observations compared to the noise-free baseline performance.
With up to 2\% noise the agent drives trajectories close to our noise-free baseline, still placing in front of the fastest human.
More noise in the agent's observations leads to jitter in its action signal, resulting in a speed loss.
With up to 9\% noise our agent is still able to drive on the track without any contact with the edge of the track.

\textbf{5) Delayed Inference:} The top of Figure~\ref{fig:delay_noise_vs_laptime} shows the lap times in setting \textit{A} when adding delays of various degrees to the agent's inference compared to the non-delayed baseline performance.
For delays up to 20~ms the agent performs close to the baseline, in some cases even slightly faster, outperforming the fastest human reference lap in each case.
With a delay of 50~ms the agent loses 0.3~seconds on the baseline due to delay related deviations from the optimal line, but still achieves a lap time among the top 10 human drivers.
With a delay of 100~ms our agent brakes too late before the sharpest curve of the track, leading to wall contact and a loss of 1.4~seconds compared to our baseline.
With a delay of 150~ms our agent is no longer able to drive a competitive racing line, often colliding with the edge of the track and losing over 6 seconds on our baseline.
It's possible that repeating the training process with the delay active is already sufficient to mitigate the decrease in performance.
In addition, a predictive model such as the one proposed in~\cite{firoiu2018human} could be applied to counter the effect of the delay by estimating the real state based on the delayed state.
In our comparison the human is disadvantaged by a relatively slow reaction time of
approximately 0.2 seconds~\cite{reactionTime}, compared to our agent, which is able to
react in approximately 4 milliseconds, including both inference and
communication with the PlayStation. We would argue, however, that this
disadvantage is part of the human system and not part of the challenge
of autonomous racing.

%% file: tables/daily_race_table.tex
    \definecolor{Lightgrey}{rgb}{0.95,0.95,0.95}
        \begin{table}[!t]
        \caption{Time Trial Comparisons Between our Approach, Human Online Competitors, and the Built-in GTS AI for the 3 Race Settings.}
        \label{tab:daily_race_results}
        \centering        
        \begin{tabular}{|l|p{20mm}|p{11mm}|p{11mm}|p{11mm}|}

            \hline
            \rowcolor{Lightgrey} Driver & Metric & Setting A & Setting B & Setting C\\

            \hline
            \textbf{Ours} & Lap time [min] & \textbf{01:15.913} & \textbf{01:39.408} & \textbf{02:06.701}\\
            
            \hline
            \multirow{3}{*}{\begin{tabular}{@{}c@{}}Human \\ players\end{tabular}} & Fastest lap [min] & 01:16.062 & 01:39.445 & 02:07.319\\ \cline{2-5}
            & Median lap [min] & 01:22.300 & 01:47.259 & 02:13.980\\ \cline{2-5}
            & \# participants & 52,303 & 28,083 & 52,335\\

            \hline
            \multirow{3}{*}{\begin{tabular}{@{}c@{}}Built-in \\ GTS \gameAI\end{tabular}} & Lap time [min] & 01:26.899 & 01:52.075 & 02:14.252\\ \cline{2-5}
            &  \begin{tabular}{@{}l@{}}Slower than x\% \\ of humans drivers\end{tabular} & 82.6\% & 80.4\% & 53.9\%\\
            \hline 
        \end{tabular}

    \end{table}

%% file: sections/conclusion.tex
\section{Conclusion}
In this paper, we have presented the first autonomous racing policy that achieves super-human performance in time trial settings in the racing simulator \textit{Gran Turismo Sport}.
The benefits of our approach are that it does not rely on human intervention, human expert data, or explicit path planning.
It leads to trajectories that are qualitatively similar to those chosen by the best human players, 
while outperforming the best known human lap times 
in all three of our reference settings, including two different cars on two different tracks.
This super-human performance has been achieved using limited computation power during both evaluation and training. 
Generating one control command with our policy network takes around 0.35~ms on a 
standard desktop PC CPU, and training, using 4~\textit{PlayStation 4} game consoles and a single desktop PC, 
took less than 73~hours to achieve super-human performance.\par

Limitations of the present work include a) the restriction to single-player time trial races without other cars on the same track, 
and b) the constrained applicability of the learned control policies to a single track / car combination.
We intend to address a) in future work by  extending the observation space to allow the perception of other cars and modify the reward to disincentivize unfair driving behavior. To extend our approach to more track / car combinations we propose using more data-efficient RL algorithms, such as meta-RL, 
where the agent can adapt to new situations with only a small amount of new samples.

%% file: sections/supplementary.tex
\section*{Appendix}

\subsection{Prior Approaches}
At an early stage of this project, we have used different learning methods, including proximal policy optimization (PPO)~\cite{schulman2017proximal} and imitation learning. Both approaches didn’t come close to the performance we have achieved with SAC. In particular, PPO required much more training data and suffered from premature convergences due to its state-independent exploration. The imitation learning method suffered from the DAgger problem~\cite{ross2011reduction} and its performance is upper bound by the training data, which was collected from human players (top 1\% driver trajectories).  We prefer to not include them due to their poor performance but rather focus more on other interesting findings, such as comparison to human players.

\subsection{GTS Access}
The GTS simulator runs on a \textit{PlayStation~4}, while our agent runs on a separate desktop computer.
We do not have direct access to the GTS simulator, neither do we have insights into the car dynamics modeled by GTS.
Instead, we interact with GTS over a dedicated API via ethernet connection.
The API provides the current state of up to 20 simulated cars and accepts car control commands, which are active until the next command is received.
Whereas previous work in the area of RL often ran thousands of simulations in parallel to collect training data~\cite{openaifive, hwangbo2017control, alphago}, we can only run one simulation 
per~\textit{PlayStation}, and for this work we make use of only 4 \textit{PlayStations} during training.
Moreover, the simulation runs in real-time and cannot be sped up for data collection 
during training nor can it be paused to create additional time for decision making.
The simulator's state is updated at a frequency of 60~Hz, but, to reduce the load on the \textit{PlayStations} when controlling 20~cars, we limit the command frequency to 10~Hz during training.
During evaluation, we switch to one car, increasing the agent's action frequency to 60~Hz.
We use a standard desktop computer with a \textit{i7-8700} processor with a clock speed of 3.20GHz for inference and a \textit{GeForce GTX 1080 Ti} graphics card for backpropagating through the models.

\subsection{Network Training} \label{sec:net_train}

To train the networks we roll out 4x20 trajectories per epoch in parallel, with a length of 100 seconds each, by using all 20 cars available on each of the 4 \textit{Playstations}.
To minimize disturbance between cars during training, we initialize the position of agents equally distributed over the racing track with an initial speed of 100 km/h, which we found can accelerate training, since it allows the agents to faster approach the maximal feasible segment speeds, which are of interest for finding the fastest possible trajectory.
    
For the training process we make use of the TensorFlow-based SAC implementation by OpenAI\footnote{\href{https://github.com/openai/spinningup/tree/b6965d89f5b713f6ad2a5fbef8f4db4fe4013250/spinup/algos/sac}{github.com/openai/spinningup}}. We modified the code base to allow to asynchronously learn during roll outs and changed the default 1-step TD error to a 5-step TD error to stabilize training as also used in~\cite{d4pg}.
The combined training of all 4 networks takes approximately 35 seconds per epoch, 
which leaves the total epoch time at 100 seconds.
Because of the real-time character of the GTS environment, an extensive hyperparameter search is not feasible.
We therefore adapt most default hyperparameters from the OpenAI implementation, except for those listed in table~\ref{tab:sac_hyperparams}.
    
\begin{table}[h]
\centering
\begin{tabular}{l|l} 
 Hyperparameter & Value\\
 \hline
 Mini-batch size & 4,096\\
 Replay buffer size & $4 \times 10^{6}$\\
 Learning rate & $3 \times 10^{-4}$\\
 Update steps per epoch & 5,120\\
 Reward scale (1/$\alpha$) & 100\\
 Exponential discount ($\gamma$) for the ``Mazda Demio" & 0.98\\
 Exponential discount ($\gamma$) for the ``Audi TT Cup" & 0.982\\
 Wall contact penalty scale ($c_{w}$)  & $5\times10^{-4}$ \\
 Number of range finders ($M$) & 13 (every 15\degree) \\
 Number of curvature mesurements ($N$) & 10 (every 0.2s) \\
\end{tabular}
\caption{Hyperparameters used for the SAC algorithm and the GTS simulation.
Denser range finders lead to slower convergence but showed no additional improvements in the learned policy.}
\label{tab:sac_hyperparams}
\end{table}



\subsection{Curvature construction}
We base the input feature selection on the results of a pre-study on GTS conducted with behavioral cloning, aiming to learn human-like driving based on a regression of expert's actions on visited states.
We iterated over combinations of subsets of features provided by GTS as well as additionally constructed features, which lead to the following feature selection:
1)~The linear velocity $\bs{v}_t \in \mathbb{R}^3$ and the linear acceleration $\bs{\dot{v}}_t \in \mathbb{R}^3$.
2)~The Euler angle $\theta_t \in (-\pi, \pi]$ between the 2D vector that defines the agent's rotation in the horizontal plane
and the unit tangent vector that is tangent to the centerline at the projection point.
This angle is the only direct way for the policy network to detect if a car is facing the wrong direction. 
3)~Distance measurements $\bs{d}_t \in \mathbb{R}^{M}$ of $M$ rangefinders that measure the distance 
from the vehicle's center point to the $M$ edge points of its surrounding objects, such as the edge of the race track.
The rangefinders are equally distributed in the front 180$\degree$ of the car's view.
4)~The previous steering command $\delta_{t-1}$. 
5)~A binary flag with $w_t=1$ indicating wall contact
and 6)~$N$ sampled curvature measurement of the course centerline in the near future $\bs{c}_t \in \mathbb{R}^{N}$.
The curvature is represented through an interpolation of the inverse radii of circles fitted through centerline points provided by GTS.
We therefore represent the observation at a given time step $t$ as a vector denoted as
$\bs{s}_t = [\bs{v}_t, \bs{\dot{v}}_t, \theta_t, \bs{d}_t, \delta_{t-1}, w_t, \bs{c}_t]$.
To allow for a fair comparison with human drivers, we only use features that humans can either directly perceive or deduce from the GTS simulation.

\begin{figure*}[!htp]
    \centering
    \includegraphics[width=1\textwidth]{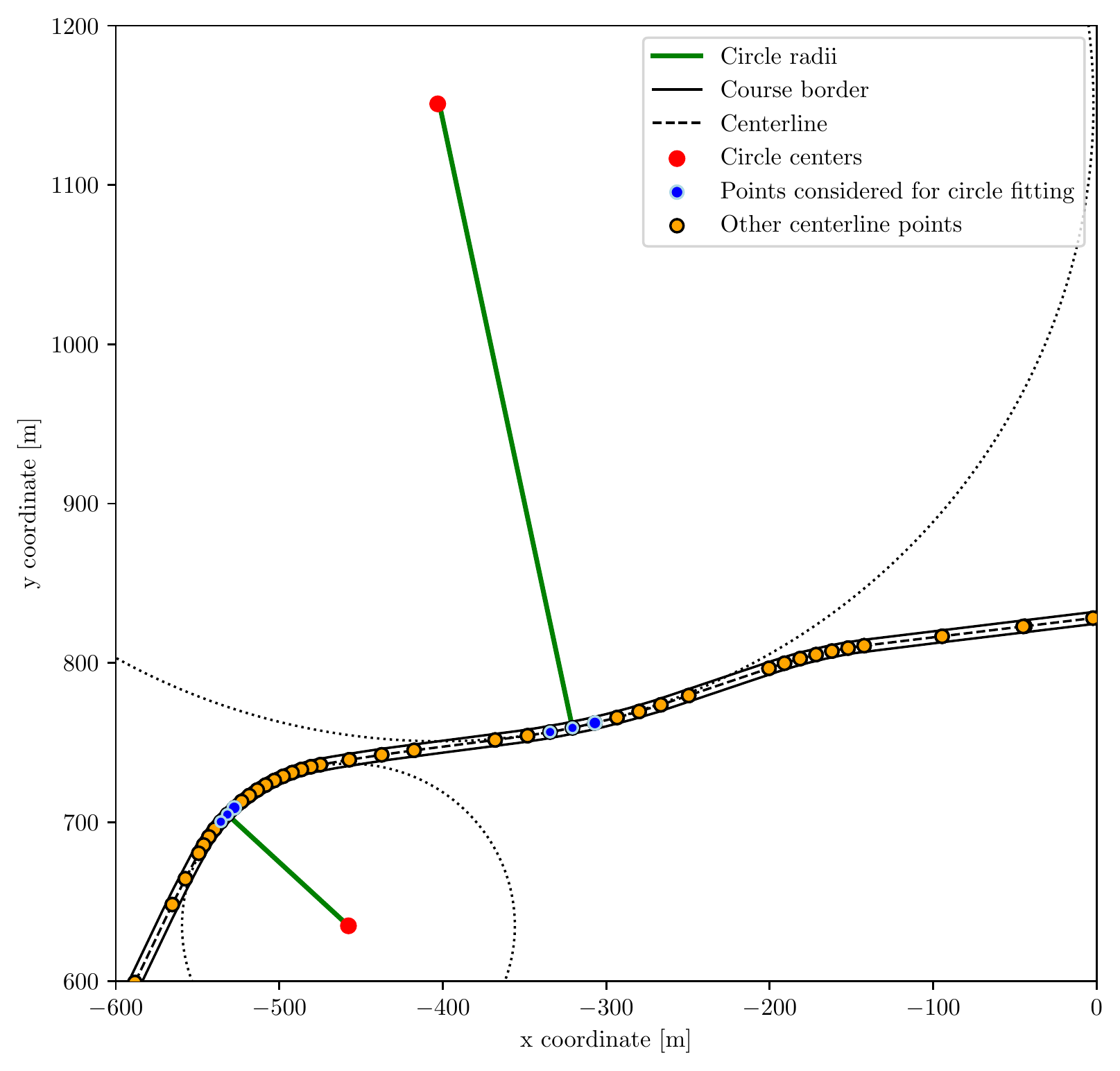}
    \caption{\textbf{Construction of the curvature measurement.} To represent the curvature of the centerline, we make use of centerline
points provided by the GTS simulation. The points are distributed such that regions with higher curvature are represented by more points. This allows getting a precise curvature measurement from looking at only three neighboring points.
We make use of that property by defining the curvature for each centerline point by the inverse radius of the circle given by that point and its two neighbors.
Right curvature is represented by negative inverted radii, left curvature by positive inverted radii.
We then interpolate between centerline points to end up with a continous representation of the curvature over the course progress.}
     \label{fig:centerline_curvature_construction}
\end{figure*}

The curvature is represented through an interpolation of the inverse radii of circles fitted through centerline points provided by GTS, as can be seen in Figure~\ref{fig:centerline_curvature_construction}. 
They are equally distributed from 1.0 to 2.8 seconds into the future from the car's current position, estimated with the car's current speed.
    
\subsection{Control Signal}
We define the action vector $\bs{a}_t = [\delta_t, \omega_t]$ by two control signals: 
a steering angle $\delta_t \in [-\pi/6, \pi/6] (rad)$ corresponding to the angle of the car's front wheels 
and a combined throttle-brake signal $\omega_t \in [-1, 1]$,
where $\omega > 0$ denotes throttle and $\omega \leq 0$ represents braking.
The magnitude of the throttle or the break is proportional to the absolute value $\|\omega\|$.
For example, $\omega=1$ is 100$\%$ throttle, $\omega=-1$ is 100$\%$ brake, 
and $\omega=0$ is no throttle and no break.
Combining the two signals reduces the complexity of the task.
Since human expert recordings show that the best human strategies do not involve 
the simultaneous use of throttle and brake, we expect not to significantly restrict the agent by combining the two signals.
To limit the outputs of the policy network to the valid range, we apply a 
\textit{tanh} activation to the output layer as proposed in~\cite{haarnoja2018soft}.

We use the automatic gearshift provided by GTS, since the used API does currently not allow 
to manually change gears.
This option is also available to human players, but experienced players mostly 
use manual gearshift to have more control of when to shift.

\subsection{Analysis of the Results by a Domain Expert}

To improve our understanding of the achieved results, we invited \textit{Gran Turismo} domain expert TG (name omitted for reasons of anonymity), who has achieved top performance in several national and international competitions, to race in our reference settings and compare his performance against our approach.
TG competed in two of our three reference settings and achieved lap times in the top 0.36 and 0.23 percentile of our human reference data.

When asked for his opinion on the policies' driving style, TG stated:
\begin{displayquote}
\emph{``The policy drives very aggressively, but I think this is only possible through its precise actions. I could technically also drive the same trajectory, but in 999 out of a 1000 cases, me trying that trajectory results in wall contact which destroys my whole lap time and I would have to start a new lap from scratch."}
\end{displayquote}

\subsection{Improvements over the built-in \gameAI}
In the 3 introduced reference settings the currently built-in \gameAI~is outperformed by a majority of human players as can be seen in Table~I of the main manuscript.
The built-in \gameAI~follows a pre-defined trajectory using a rule-based tracking approach, similar to other trajectory following approaches that have been widely studied in the control community~\cite{hellstrom2006follow,ritzer2015advanced,ni2019robust}.
Due to the non-linearity of the GTS dynamics, a slight deviation from such a trajectory strongly changes the new optimal trajectory, which makes it practically impossible to pre-compute and then track an optimal trajectory.
The built-in \gameAI~solves this problem by using a cautious reference trajectory which includes a margin to the track's borders and allows recovery when deviating from the trajectory, as can be seen in Figure~8 of the main manuscript.
This helps reducing the risk of contacting the track's side walls when deviating from the trajectory.
However, it also leads to the trajectory's curves having smaller radii, forcing the \gameAI~to decelerate to not lose traction, resulting in curve exit speeds up to 50 km/h slower than those of our approach and the fastest human.
The bottom of Figure~8 of the main manuscript shows how the built-in \gameAI~brakes in segments where neither our approach nor the fastest human brake, to be able to follow its reference trajectory.
By learning a flexible driving policy without any explicit restrictions, our approach was able to overcome the shortcomings of the previous built-in \gameAI~and learn a driving policy more similar to that of a human expert driver regarding speed and path.

\subsection{Reproducibility of experiments after GTS Update 1.57}\label{sec:reproducibility}

Starting with {\color{blue}\href{https://www.gran-turismo.com/us/gtsport/news/00_5762273.html}{GTS version 1.57}} released on April 23rd 2020, the simulation dynamics for the ``Audi TT Cup '16'' changed such that the achievable times for setting \textit{A} and \textit{C} of this paper are different.
Under the new update a fair comparison to the human data sets of setting \textit{A} and \textit{C}, which were collected under an older GTS version, is therefore no longer possible.
For future comparison with this work we recommend the following human data set as a replacement for setting \textit{A}: {\color{blue}\href{https://www.kudosprime.com/gts/rankings.php?sec=daily&eid=22164,22170,22167}{kudosprime.com}}.
The simulation settings used for this race are equal to those used in setting \textit{A} of this paper, except that they're using the updated car dynamics from version 1.57+ as well as updated tires (RM instead of RH).
The approach described in this paper achieves a lap time of 74.686 seconds after 5 days of training when applied on this new setting, as shown in Table~\ref{tab:update_157_results}.
    \definecolor{Lightgrey}{rgb}{0.95,0.95,0.95}
        \begin{table}[!t]
        \caption{Time Trial Comparisons Between our Approach, Human Online Competitors, and the Built-in GTS AI for the new setting \textit{A} under GTS version 1.57.}
        \label{tab:update_157_results}
        \centering        
        \begin{tabular}{|l|p{40mm}|p{11mm}|}

            \hline
            \rowcolor{Lightgrey} Driver & Metric & Updated setting \textit{A}\\

            \hline
            \textbf{Ours} & Lap time [min] & \textbf{01:14.686}\\
            
            \hline
            \multirow{3}{*}{\begin{tabular}{@{}c@{}}Human \\ players\end{tabular}} & Fastest lap [min] & 01:14.775\\ \cline{2-3}
            & Median lap [min] & 01:21.794\\ \cline{2-3}
            & \# participants & 71,005\\

            \hline
            \multirow{2}{*}{\begin{tabular}{@{}c@{}}Built-in \\ GTS \gameAI\end{tabular}} & Lap time [min] & 01:26.356\\ \cline{2-3}
            &  Slower than x\% of humans drivers & 82.5\%\\
            \hline 
        \end{tabular}

    \end{table}